\documentclass[conference]{IEEEtran}
\IEEEoverridecommandlockouts
\usepackage{cite}
\usepackage{amsmath,amssymb,amsfonts}
\usepackage{algorithmic}
\usepackage{enumerate}
\usepackage{graphicx}
\usepackage{hyperref}
\usepackage{textcomp}
\usepackage{xcolor}
\usepackage{mwe}
\usepackage{caption,subcaption}
\DeclareMathOperator*{\argmin}{argmin} 

\hypersetup{
    colorlinks=true,
    linkcolor=blue,
    filecolor=magenta,      
    urlcolor=cyan,
}

\newcommand{\Softbubble}{\emph{Soft-bubble} } 
\newcommand{\Softbubbles}{\emph{Soft-bubbles} }
\newcommand{\softbubble}{\emph{Soft-bubble} }
\newcommand{\softbubbles}{\emph{Soft-bubbles} }

\newif\ifcomments
\commentstrue 

\def\BibTeX{{\rm B\kern-.05em{\sc i\kern-.025em b}\kern-.08em
    T\kern-.1667em\lower.7ex\hbox{E}\kern-.125emX}}
    
\begin{document}

\title{\LARGE \bf
 \softbubble grippers for robust and perceptive manipulation
}

\author{Naveen Kuppuswamy, Alex Alspach, Avinash Uttamchandani, Sam Creasey, Takuya Ikeda, and Russ Tedrake$^{*}$
\thanks{$^{*}$ All authors are with the Toyota Research Institute, One Kendall Square, Building 600, Cambridge, MA 02139, USA, {\tt\small [firstname.lastname]@tri.global}}%
}

\maketitle

\begin{abstract}
Manipulation in cluttered environments like homes requires stable grasps, precise placement and robustness against external contact. We present the \Softbubble gripper system with a highly compliant gripping surface and dense-geometry visuotactile sensing, capable of multiple kinds of tactile perception. We first present various mechanical design advances and a fabrication technique to deposit custom patterns to the internal surface of the sensor that enable tracking of shear-induced displacement of the manipuland. The depth maps output by the internal imaging sensor are used in an in-hand  \emph{proximity} pose estimation framework - the method better captures distances to corners or edges on the manipuland geometry. We also extend our previous work on tactile classification and integrate the system within a robust manipulation pipeline for cluttered home environments. The capabilities of the proposed system are demonstrated through robust execution multiple real-world manipulation tasks. A video of the system in action can be found \href{https://youtu.be/G_wBsbQyBfc}{here}.
\end{abstract}


\section{Introduction}



Incorporating mechanical compliance, or softness, into manipulators has been observed to be beneficial in making grasping more robust \cite{Hughes2016}. However, many in-home tasks necessitate not just robust picking, but precise placement, e.g. dishwasher loading, liquid container handling, and shelf stacking. Although compliance can somewhat mitigate inaccuracies in pre-grasp pose estimation and object shape variations, variability in the post-grasp state may adversely impact task success. This problem is further exacerbated in highly cluttered home environments with visual occlusions, tight spatial constraints, and unforeseen contacts. Augmenting compliant end effectors with tactile perception capabilities might help compensate for variability in the post-grasp state of manipulands and for unexpected disturbances. To address these challenges, we present the highly compliant \softbubble gripper system which integrates multiple tactile perception capabilities in order to enable robust manipulation in tightly constrained environments.


While progress has been made in tactile sensing techniques, open questions remain on the ideal design for soft or compliant surfaces as well as on what kind of perception they should enable. For manipulation in cluttered environments, tactile sensing should enable at least a few of the following capabilities: manipuland pose estimation (to estimate post-grasp pose), manipuland external force estimation (to estimate both expected and unexpected forces), slip detection, object classification (e.g. to compensate for occlusion when grasping in clutter), and grasp quality estimation. This paper tackles several of these problems.


\begin{figure}[ht]
    \centering 
    \includegraphics[width=1.0\columnwidth]{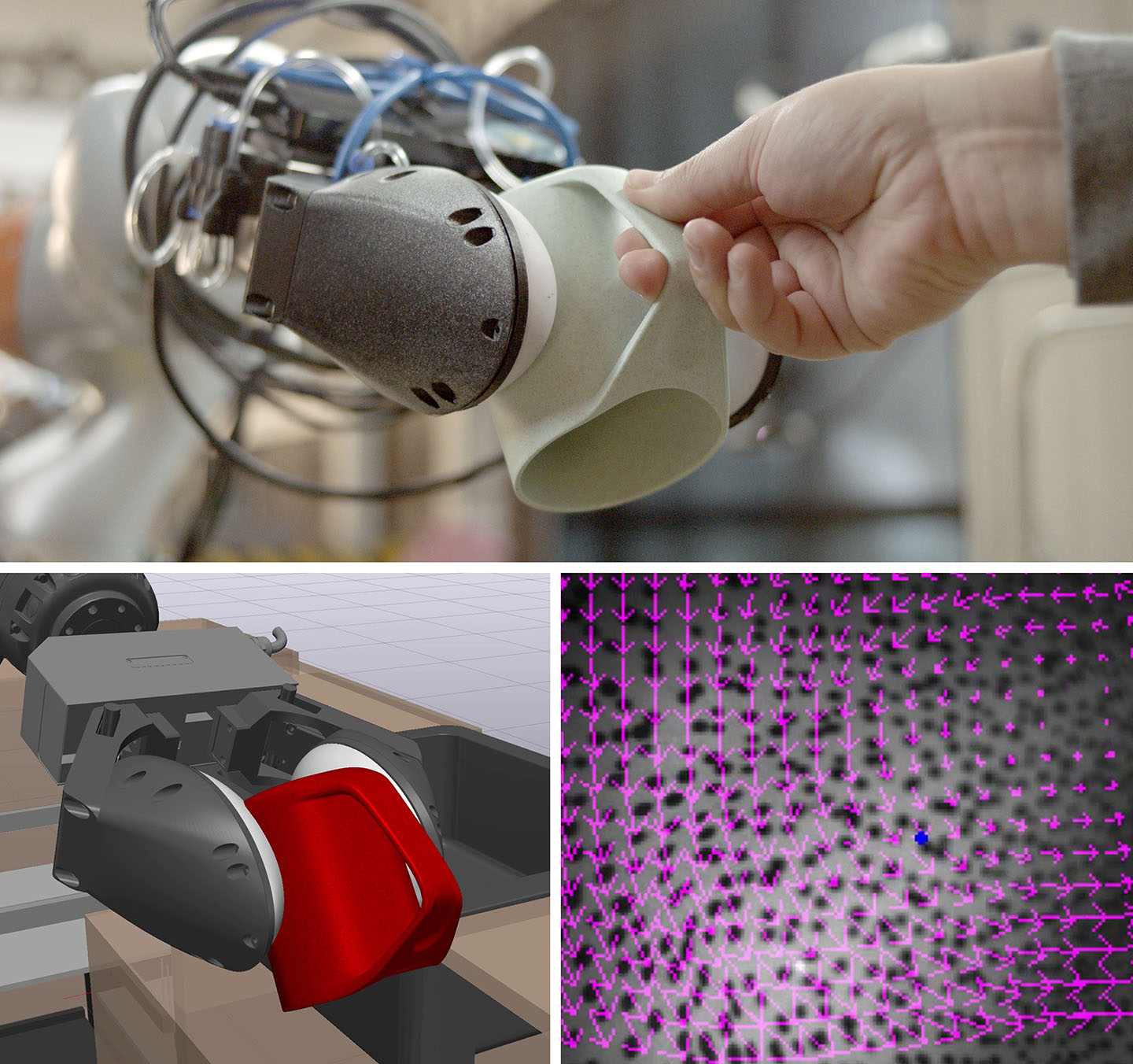}
    \caption{Top: Highly compliant \softbubble parallel gripper on a robot arm grasping an object. Bottom-left: In-hand pose estimate during this interaction; Bottom-right: Shear-induced displacements tracked on the interior of this sensor.}
    \label{fig:soft_bubble_schunk}
\end{figure}

The \softbubbles \cite{Alspach2019, kuppuswamy2019fast} combine a highly compliant manipulation surface with dense geometry sensing. Mechanically, they achieve robust grasps since they are highly deformable, are easy to build due to their air-filled membrane design, and are fairly durable. They are also closely related to other visuotactile sensors reported in literature \cite{Shimonomura2019}, with the key difference that the generated depth maps are directly measured by the internal imaging sensor and are not inferred. Some notable limitations on the prior iterations of the \softbubbles were their large form factor and an inability to track shear forces on the contact surface. This meant they could neither be used for human-scale object manipulation nor compensate for externally induced shear. Meaningfully co-designing the mechanical and perceptual aspects of these sensors for a robust manipulation pipeline was an open question - one which is directly tackled in this paper.

We present four key contributions to the \softbubble technology that combine to allow for robust manipulation of real-world, human-scale objects in a cluttered home environment: 

\begin{enumerate}[(A)]
    \item \label{contrib:gripper} Design and fabrication techniques for a smaller parallel gripper form factor, leveraging modifications of the previously used time-of-flight (ToF) camera system
    \item \label{contrib:shear} Techniques for adding high-density internal markers to the bubble surface that enable dense optical flow pattern tracking for estimating shear-induced membrane displacement
    \item \label{contrib:pose} An optimization-based \emph{proximity} pose estimation framework that is inspired by current state-of-the-art depth tracking methods.
    \item \label{contrib:classification} Integrated tactile classification, using simultaneous images and an automated ground-truth labeling pipeline
\end{enumerate} 

We demonstrate the capabilities of the \softbubble gripper system through various tasks that test its sensing effectiveness, robustness to uncertainty, as well as its efficacy in human-robot interaction and antagonism. We conclude with observations on the performance, the gripper's limitations, and a discussion of future work.

\section{Related Work}
\label{sec:related_work}
The contributions presented in this paper cover four key areas and the related work for each is summarized in this section.
\subsection{Compliant Gripper Design}
In contrast to most compliant gripper designs, which are based on solid materials \cite{Hughes2016}, the \softbubbles use an air-filled deformable structure such as those reported in Kim et al \cite{Kim2015}. They are also a high-resolution visuotactile sensor, like GelSight \cite{Yuan2017, wilson_design_2020}, and GelSlim \cite{Donlon2018}, which is advantageous for precision manipulation. The air-inflated latex membrane structure of the \softbubble has multiple mechanical benefits. (i) a large degree of compliance through a combination of pneumatic effects and membrane elasticity\footnote{Typically, the deformations exhibited by \softbubbles are at least an order of magnitude greater than GelSight or GelSlim (multiple cm vs multiple mm on human-scale objects like mugs)}, (ii) the resulting higher friction at the contact patch leads to better grasps, (iii) they are potentially more robust to wear-and-tear from long-term use. 

\subsection{Shear Estimation}
Estimating shear and/or slip from visuotactile sensing has been investigated \cite{yuan2018}, particularly for ensuring stable and sensitive grasps that are robust to unforeseen external contacts. A popular approach has been to embed uniform visual markers either on the contacting surface or within the the sub-surface material. Methods have also been proposed for decomposing this shear information into tangential and torsional components \cite{zhang2019}.


However, due to the large deformations and ellipsoidal form factor of the \softbubbles, such uniform markers result in shear tracking errors. Therefore we developed a pseudorandom pattern deposition technique that enables dense optical flow processing for shear-deformation tracking.

\subsection{In-hand Pose Estimation}
The problem of estimating the post-grasp pose of a manipuland within an end effector's grasp has been attempted both with and without the use of tactile sensors. State-of-the-art methods for tracking known object geometries using depth sensors typically leverage probabilistic filters that account for unforeseen occlusions \cite{issac2016depth}. One application of tactile sensing has been to improve the stability of the filter under contact by collapsing the state to a contact manifold \cite{koval2013pose}. Algorithms such as DART overcome some of the intractability issues in filter-based tracking by solving an optimization problem and can also deal with articulated objects \cite{schmidt2014dart}; more accurate extensions have been proposed that account for physical non-penetration constraints when in contact with known surfaces \cite{schmidt2015depth}. 


In the case of visuotactile sensors, the rendered depth maps can be of sufficiently high resolution to be used directly for in-hand pose estimation \cite{li2014localization}. Direct depth registration methods such as Iterative Closest Point (ICP) can be employed for local pose estimation \cite{Alspach2019} given a sufficiently close initial guess that can either be learned \cite{bauza2019tactile} or be obtained from methods that can compute the contact-patch resulting from contact with a tactile sensor \cite{kuppuswamy2019fast}. Also, external depth sensor information can be fused with tactile depth \cite{Izatt2017}.

The \emph{proximity} pose estimator we present in this paper is closely related to optimization-based methods like DART using visuotactile sensors \cite{Izatt2017} due to their robustness against initialization - we propose some strategies to improve the smoothness of the cost function. The deformations of the \softbubbles under stable grasp capture enough of the object geometry to enable an optimization-based pose estimator from tactile information alone, as presented in this paper.

\subsection{Tactile Classification}
Tactile material and object classification is an interesting use case for tactile sensors \cite{liu2017} since it enables recognition of the object that is being grasped even when it is occluded from other vision sensing. As in \cite{Alspach2019}, we use a ResNet18 architecture \cite{he2016deep} image classifier; we now train and infer on concatenated images simultaneously combining individual depth or IR images from each of the sensors. An automatic labelling pipeline was developed for generating sufficient quantities of labelled images. 

\section{Mechanical Design} 
\label{sec:mech_design}
\begin{figure}[ht]
    \centering 
    \includegraphics[width=0.98\columnwidth]{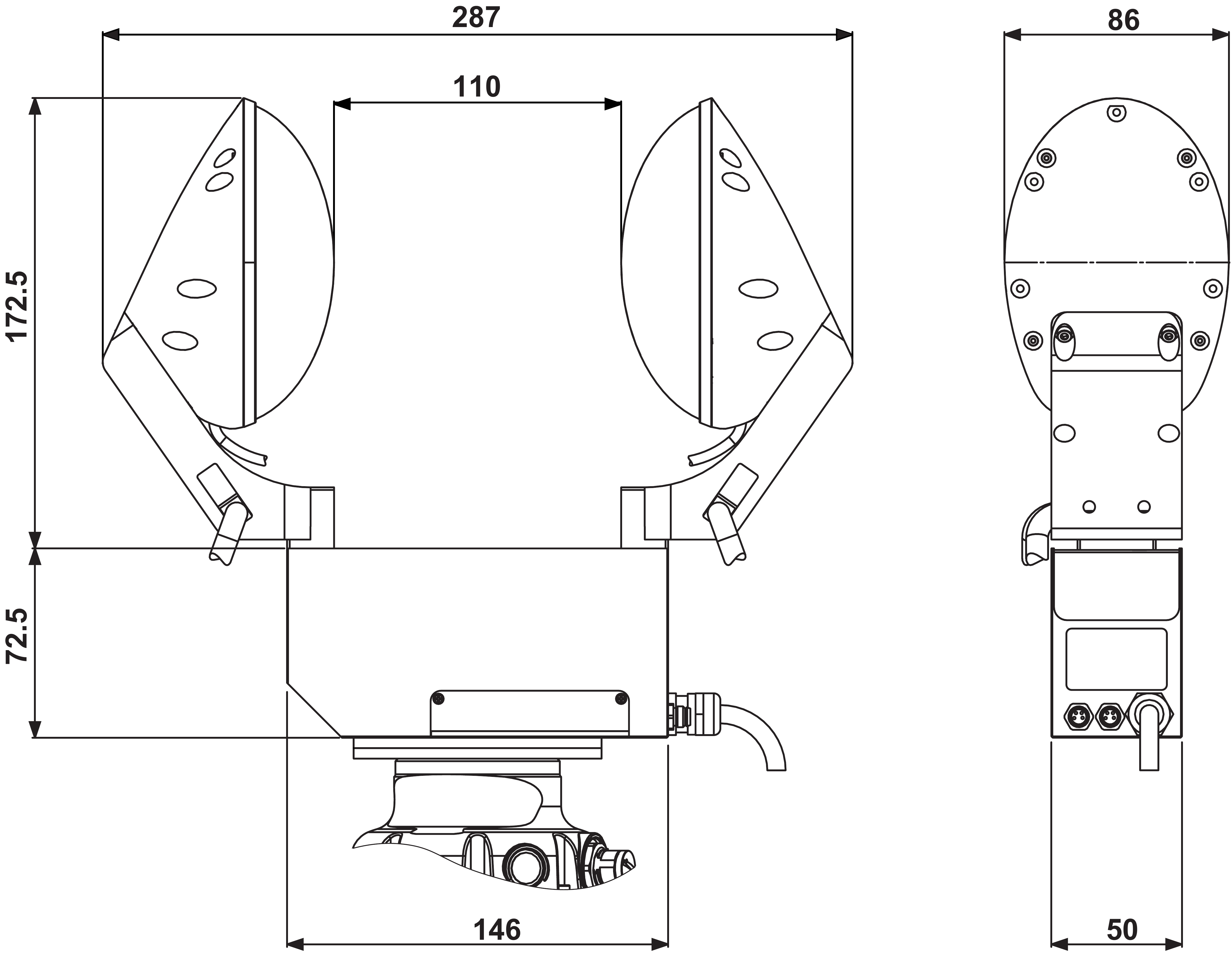}
    \caption{The \softbubble gripper is designed to manipulate household objects in tight spaces. All dimensions in mm.}
    \label{fig:bubble_gripper_design}
\end{figure}

Our previous \softbubble prototype was relatively large with a single sensor used as an end effector \cite{Alspach2019}. The new \softbubble fingers, as seen in Fig.~\ref{fig:bubble_gripper_design}, have been designed to both task-based and perceptual requirements. In order to achieve tasks in constrained domestic environments, the \softbubbles are designed to be attached to a standard parallel gripper, to interact with human-scale objects, and to fit into tight household spaces (e.g. a sink or dish washer). Perceptually driven improvements include the use of a shorter range depth sensor, interchangeable ``bubble modules,'' as well as a method for adding visually trackable patterns to a bubble's internal surface.

\subsection{Internal Depth Sensor}

\begin{figure}[ht]
    \centering 
    \includegraphics[width=0.98\columnwidth]{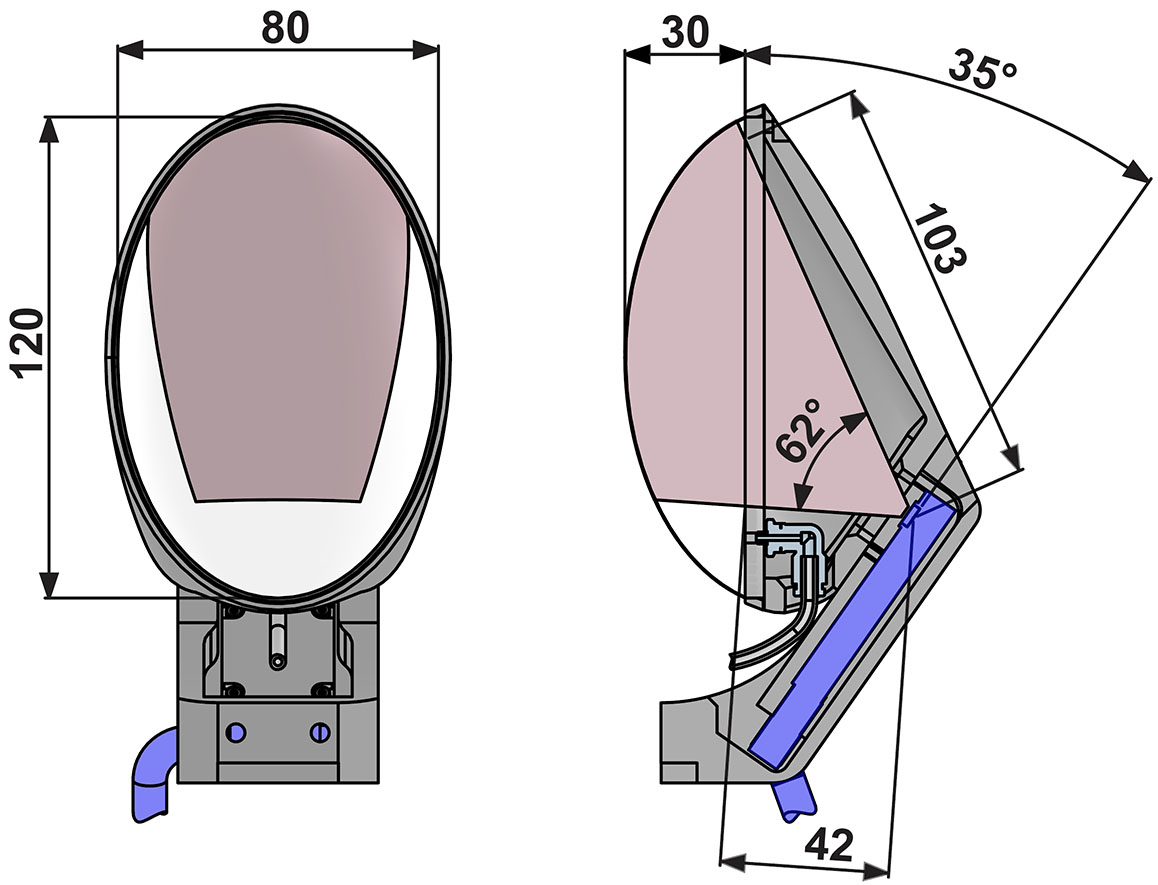}
    \caption{An angled, short-range ToF depth sensor (blue) maximizes FOV (red) and depth measurement accuracy while minimizing overall gripper width. All dimensions in mm.}
    \label{fig:bubble_finger_design}
\end{figure}

For shorter-range depth sensing (as compared to the off-the-shelf picoflexx), a prototype ToF depth sensor from PMD with a shorter imager-emitter baseline was used. This shorter baseline allows for more consistent illumination of the deformable bubble membrane at closer distances, and therefore permits more accurate depth measurements. As seen in Fig.~\ref{fig:bubble_finger_design}, the internal ToF sensor is angled relative to the contact surface, with a working range of 4-11~cm. Depth measurements are more accurate near the upper bound of this range. The angled ToF depth sensor maximizes field of view (FOV) and depth measurement accuracy at the center and tips of the fingers where contact happens most, while minimizing the overall gripper width.

\subsection{Bubble Modules}



\begin{figure}[ht]
    \centering 
    \includegraphics[width=0.80\columnwidth]{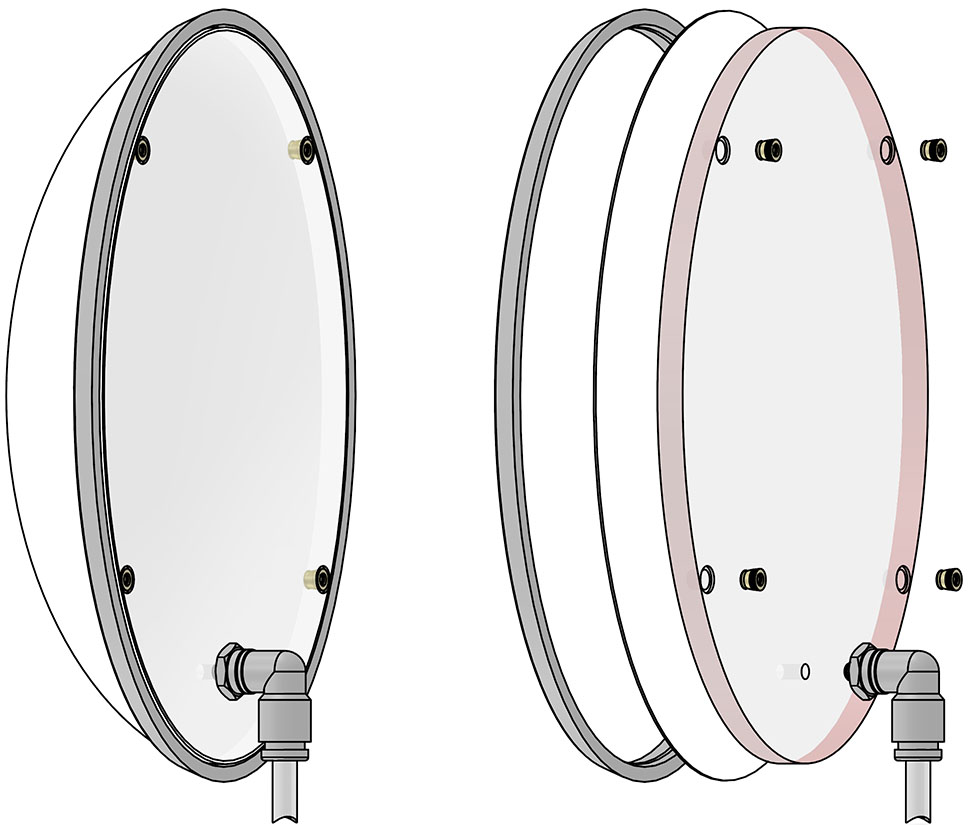}
    \caption{Assembled and exploded views of bubble module components including latex membrane, clear acrylic plate, ring that mounts the membrane onto the acrylic, threaded inserts, tube fitting and tube. The red surface is where CA glue is applied during assembly to seal the module.}
    \label{fig:bubble_cartridge}
\end{figure}

A modular design allows for quick swapping of bubbles. Bubbles may be exchanged for various reasons including the replacement of a damaged module, switching mechanical properties like membrane thickness, or swapping internal patterning based on the needs of the perception system. The modules consist of a plain or patterned latex membrane, a clear acrylic backing plate, a ring that mounts the membrane onto the acrylic, threaded inserts, and a tube fitting for inflation and pressure sensing. An exploded view of this cartridge can be seen in Fig.~\ref{fig:bubble_cartridge}. The components are sealed together using CA glue. The sensor electronics remain mounted to the gripper, while the bubble cartridge can be removed and replaced using four screws.


\subsection{Printed Patterning}
\label{sec:shear_pattern}

\begin{figure}[ht]
    \centering 
    \includegraphics[width=0.98\columnwidth]{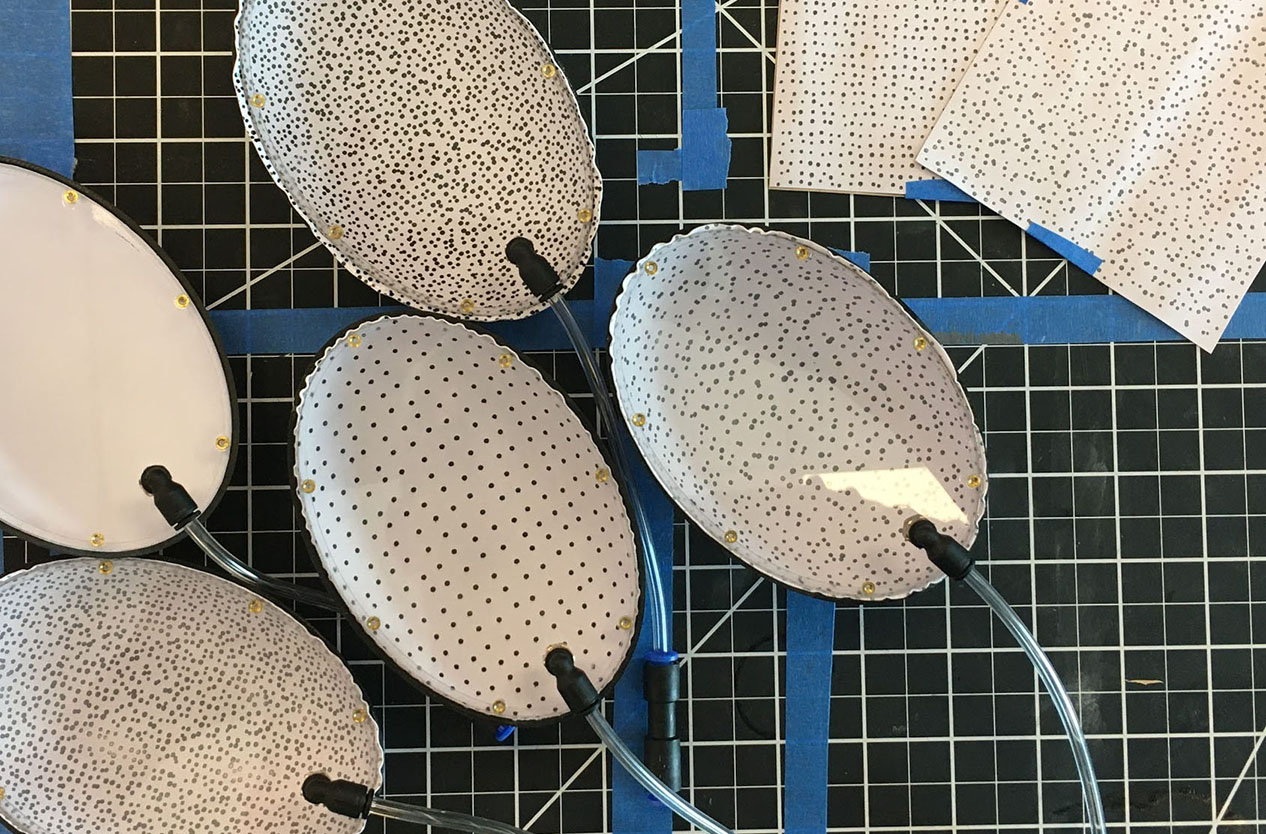}
    \caption{Experimental bubble modules and stencils with a variety of pseudorandom printed internal dot patterns for shear tracking, with dots in both black and gray.}
    \label{fig:shear_patterns}
\end{figure}

The various patterns seen in Fig.~\ref{fig:shear_patterns} are script-generated with parameters controlling dot density (dots/mm$^2$), minimum and maximum dot diameter, and pattern randomness. These patterns are output as vector files for laser cutting adhesive-backed stencils. These stencils are adhered to plain sheet latex, a thin layer of balloon screen printing ink is painted over the stencil, then the stencil is removed while the ink is still wet. This method can be used to apply patterns of single and multiple colors.



\subsection{Construction and Durability}

A batch of \emph{Bubble-sensor} fingers can be inexpensively assembled in as little as two hours with no more than a FDM printer, laser cutter, scissors, glue and a paint brush; there is no casting, as we use off-the-shelf sheet latex for the membrane. In long-running experiments, we were able to operate the \softbubbles for over 100 hours before the first failure - a puncture from the sharp edge of a manipuland. This type of failure is simple to recover from. When a puncture occurs, the surrounding area is lightly sanded then cleaned. A small patch is cut from latex, then both the patch and area surrounding the puncture are thinly painted with rubber cement, allowed to dry for a few seconds, then the patch applied. After a few minutes, the bubble can be re-inflated with negligible effect on the associated tactile sensing capabilities. In the event that a bubble can not be patched, the module is then swapped out for a backup bubble. We are currently exploring techniques to improve the durability and the mean time to failure.

\section{Proximity Pose Estimation}
\label{sec:pose_estimation}
As mentioned in Sec. \ref{sec:related_work}, our approach to pose estimation of known objects is closely related to optimization-based methods for tracking on point-clouds \cite{schmidt2014dart, schmidt2015depth, Izatt2017} with a few modifications. In our prior work \cite{Alspach2019, kuppuswamy2019fast}, we observed that quality and resolution of the pair of images produced by the \Softbubbles on contact are more than sufficient to enable tracking of manipuland pose. Here, we focus on a method that is more robust to estimator initialization - this is vital in the case of manipulating in cluttered environments where a priori guesses on in-hand state may be poor or non-existent.

\begin{figure}[ht]
\centering 
  \includegraphics[width=0.7\linewidth]{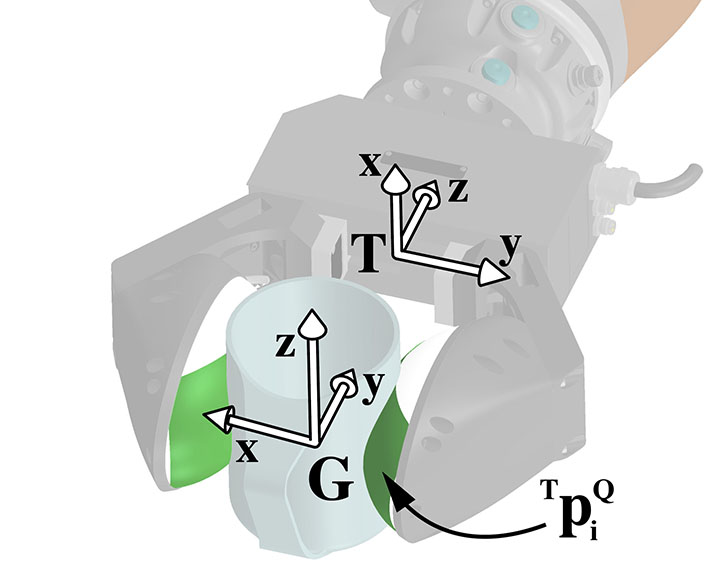}
\caption{The end-effector and manipuland geometry frames for the in-hand pose estimation problem. The shaded region in green represents the point-cloud corresponding to each of the \Softbubble sensors.}
\label{fig:in_hand_pose_problem}
\end{figure}

Consider the scenario presented in Fig.~\ref{fig:in_hand_pose_problem}. There is a rigid object $O$ with an associated geometry reference frame $G$ that is held between two \softbubbles mounted on the fingers of a robot gripper. There is a frame $C_i$ associated with each of the \softbubbles and the gripper in turn has a tool frame $T$ associated with it. The single-shot in-hand pose estimation problem is to estimate ${^G}X_{T}(t)$, for a known object geometry (where ${^G}X_{T} \in \mathrm{SE}3$), when the object is under a stable grasp (the object is not slipping) using available sensor measurements $Y(t)$. 

As previously suggested by Schmidt et al\cite{schmidt2014dart}, signed distance functions (SDF) are a convenient parameterization of the object geometry since they can overcome the difficulty in finding point-pair correspondences between the reference geometry and the sensed point-clouds. However, the non-smoothness of the SDF gradients on geometries with sharp edges or corners can prove detrimental to the optimization. 

\begin{figure}[ht]
    \centering 
\begin{subfigure}{.48\columnwidth}
  \includegraphics[width=\linewidth]{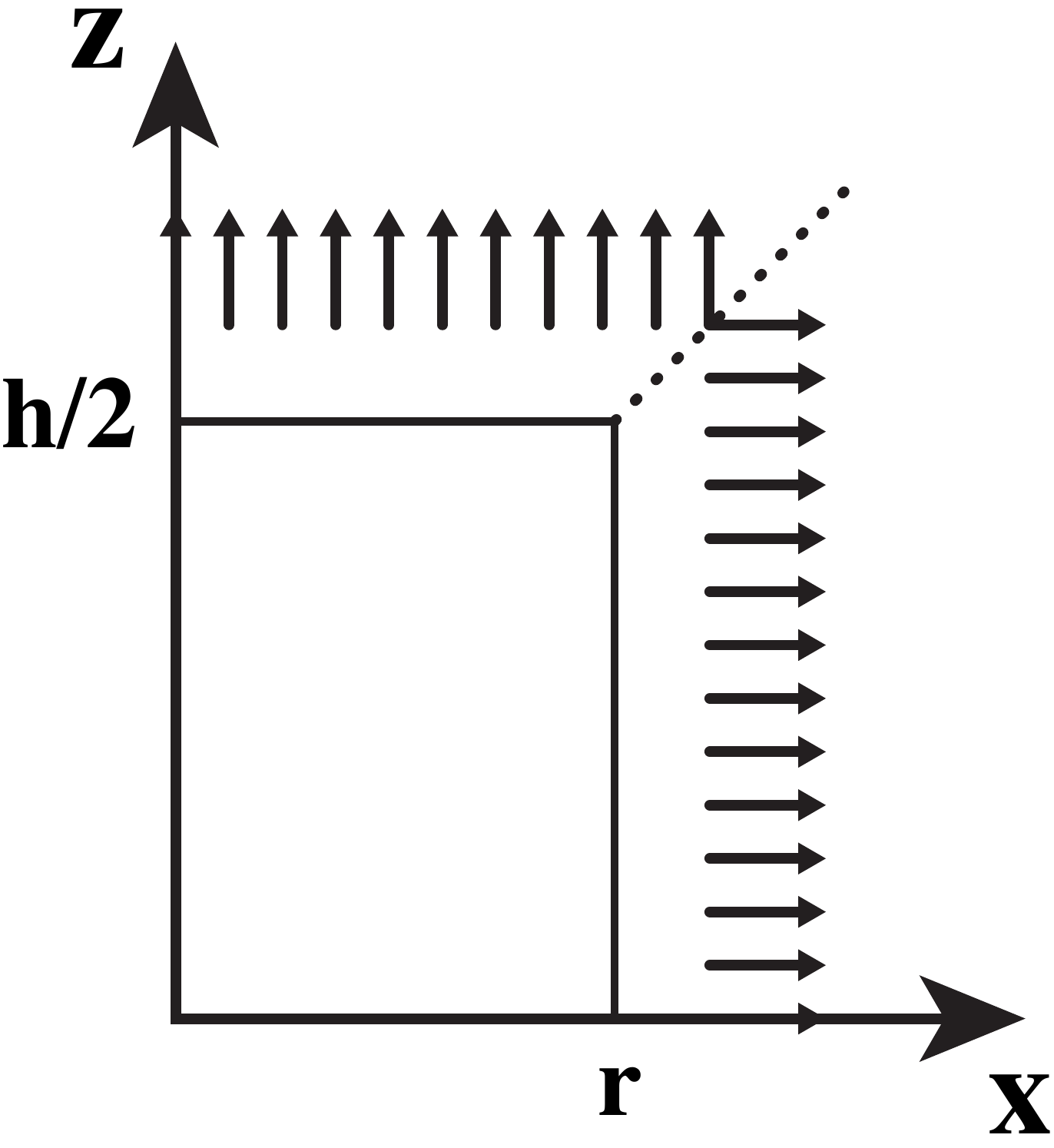}
  \caption{}
  \label{fig:corner_correction:1}
\end{subfigure}\hfil 
\begin{subfigure}{.48\columnwidth}
  \includegraphics[width=\linewidth]{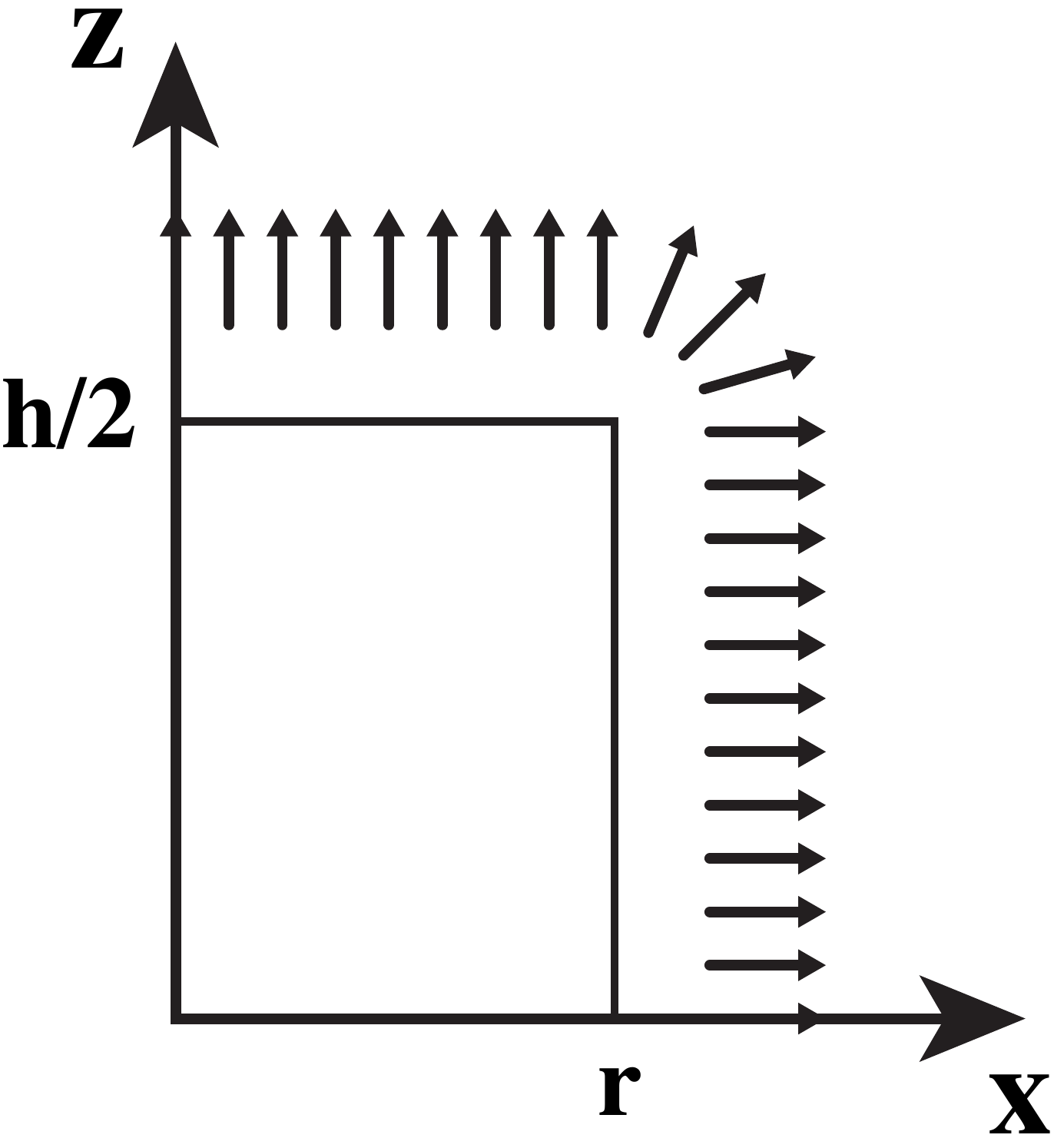}
  \caption{}
  \label{fig:corner_correction:2}
\end{subfigure}\hfil 

\caption{Proximity field gradients illustrated on a 2D projection of a cylinder geometry; (a) signed-distance fields without corner correction and (b) proximity fields with corner correction.}
\label{fig:corner_correction}
\end{figure}

Therefore, we extend the notion of the SDF to a more generic scalar \emph{proximity} field that exhibits smooth gradients outside a given geometry even when sharp corners or edges are present. This  can be defined as the function $\phi$, a scalar field that is specified for a given geometry with an associated reference frame $G$ that maps any point P that is specified in a frame $C$ to a scalar value $c$ as in,

\begin{equation}
    \phi({^G}X^C{^C}p^P) = c, {^C}p^P \in \mathrm{R}^3, \quad c \in \mathrm{R}^1,
    \label{eq:proximity_field}
\end{equation}
where the notation ${^C}p^P$ denotes a position vector p between C and P, and ${^G}X^C$ denotes the transform between $C$ and $G$. By convention $c\ge0$ outside the geometry, $c<0$ inside and $c=0$ at the boundary. It is desirable that the gradients for a proximity field are well-defined and smooth for all ${^G}p^P \in \mathrm{R}^3$. For smooth surfaces, similar to SDFs, the gradient satisfies the Eikonal equation $\nabla \phi = 1$. However, it is important that the smoothness is also guaranteed outside a geometry which can be ensured by considering the nearest corner distance along with the nearest surface distance as depicted in Fig. \ref{fig:corner_correction} for the cylinder case.

Then, given a point-cloud measurement $^Tp{^Q}_{{set}_i} = [^Tp{^Q}_0, \dots, ^Tp{^Q}_N]$ that is specified in the end effector frame $T$, the problem in its simplest form can be posed as a nonlinear optimization given by,

\begin{equation}
\begin{split}
\theta^* = &\argmin_{\theta} \sum_{i=0}^{N} \left[ \phi({^G}X_T(\theta)\; ^T{p_i}{^Q}) \right]^2, \quad  \text{s.t.}: \theta \in \mathrm{SE}(3),
\end{split}
\label{eq:proximity_pose_optimization}
\end{equation}
where $\theta$ is the chosen parameterization of the pose ${^G}X_T$, for a $7$-dimensional vector composed of the three translational coordinates and four rotational coordinates (represented by the quaternion). We use a unit-norm constraint on the the quaternion part of $\theta$.

Here, we specifically focus on analytic \emph{proximity} fields for cylinders which we develop to ensure smoothness of $\nabla \phi$ outside the geometry. This function can be defined in the following manner:

Given a cylindrical geometry of height $h$ and radius $r$, with the height along the z-axis, and the half-height at the origin of G, consider a slab $\delta_{sl}$ consisting of the planar projection of the cylinder, then the \emph{proximity} field $\phi$ can be defined as,

\begin{equation}
    \phi({^G}p^P) = \|max(\delta{_{sl}}, 0_2) \|_2 + vmax(min(\delta{_sl}, 0_2)),
    \label{eq:cylinder_proximity_field}
\end{equation}
where the cylinder slab function $\delta{_{sl}}$ is defined by,
\begin{equation*}
    \begin{split}
    \delta{_{sl}} &= |d| - b, \\ 
    d &= [\|{^G}p^P(x, y)\|_2, {^G}p^P(z)]^T,  \\
    b &= [r, h/2]^T,
    \end{split}
\end{equation*}
where the operators $max$ and $min$ are the binary operators returning the maximum and minimum of a pair of scalar values and $vmax$ denotes the operator that return the $max$ value among the components of a given vector, $x$, $y$, and $z$ denote the components of ${^G}p^P$ and the modulus operation $|d|$ is element-wise. Fig. \ref{fig:corner_correction:2} illustrates the field at some points outside the cylinder.

This algorithm possesses several advantages from the perspective of implementation and efficiency. First, the analytical form of the \emph{proximity} fields for several basic primitive shapes can be easily derived and thus efficiently computed. Second, similar to SDFs, multiple primitive \emph{proximity} fields can be aggregated for complex geometries and the gradients can be analytically computed. We use automatic differentiation to efficiently compute the gradients, and solve the constrained nonlinear optimization in Eq. \ref{eq:proximity_pose_optimization} using SNOPT \cite{gill2005snopt}. 

\begin{figure}[t]
    \centering 
\begin{subfigure}{0.48\columnwidth}
  \includegraphics[width=\linewidth]{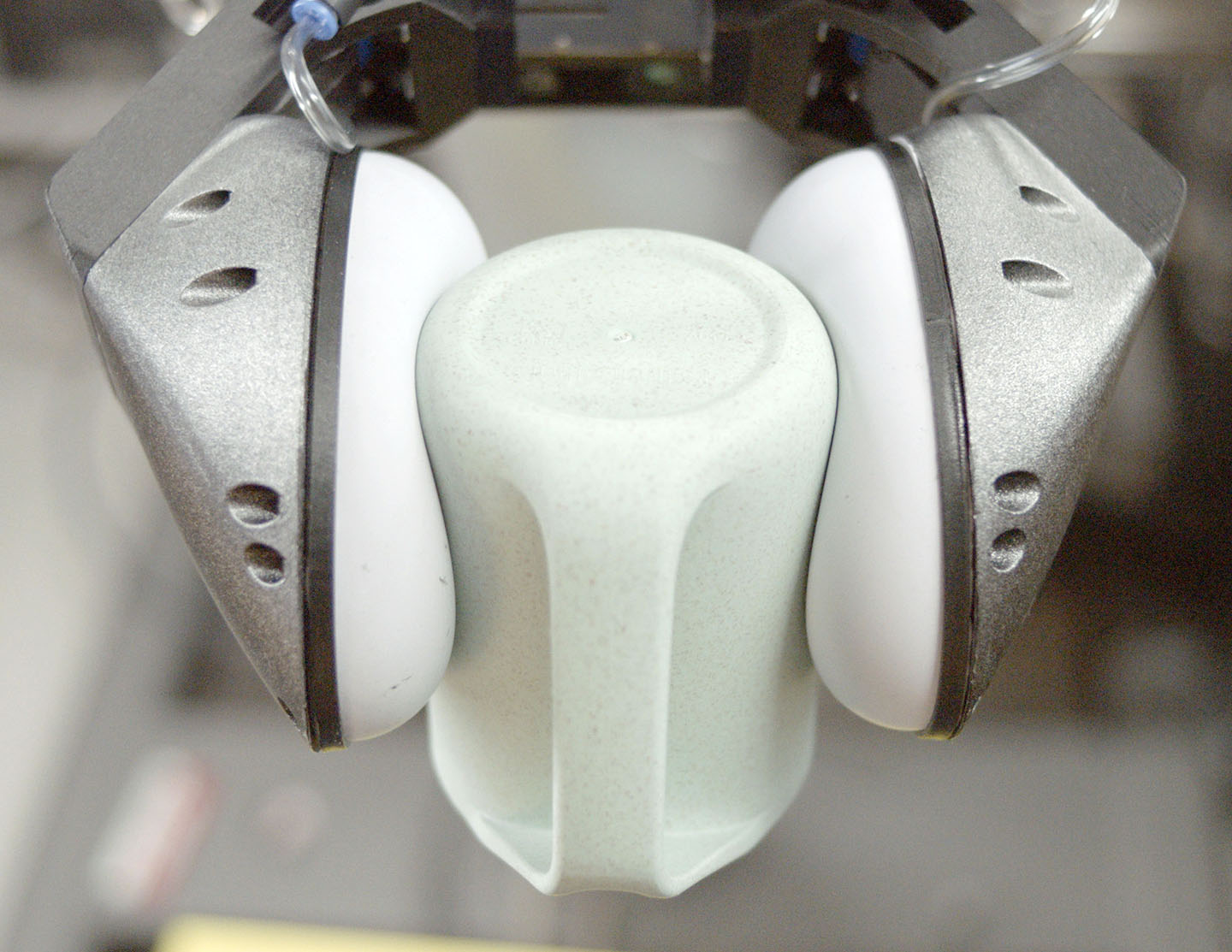}
  \caption{}
  \label{fig:proximity_pose_optimizer:1}
\end{subfigure}\hfil 
\begin{subfigure}{0.48\columnwidth}
  \includegraphics[width=\linewidth]{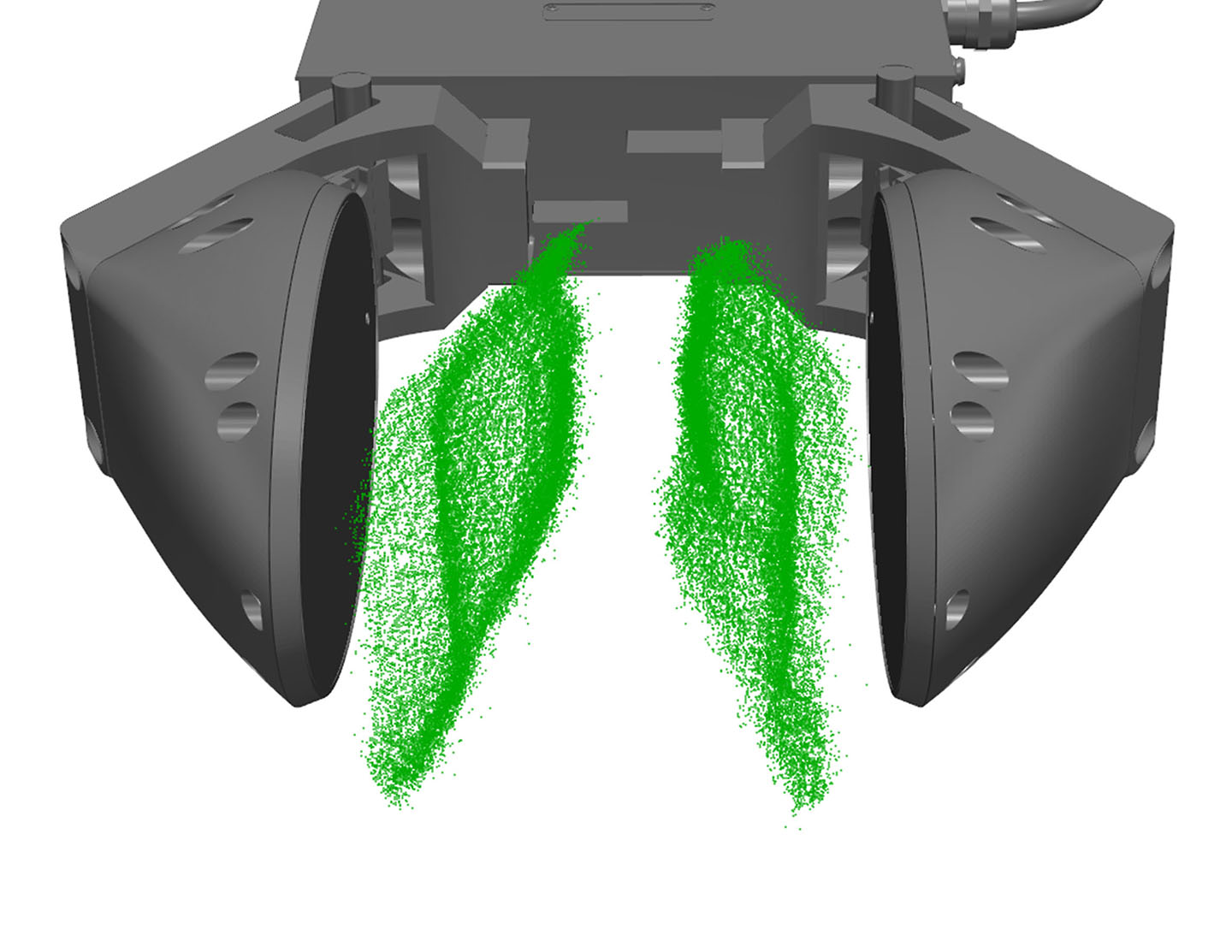}
  \caption{}
  \label{fig:proximity_pose_optimizer:2}
\end{subfigure}\hfil 

\medskip

\begin{subfigure}{0.48\columnwidth}
  \includegraphics[width=\linewidth]{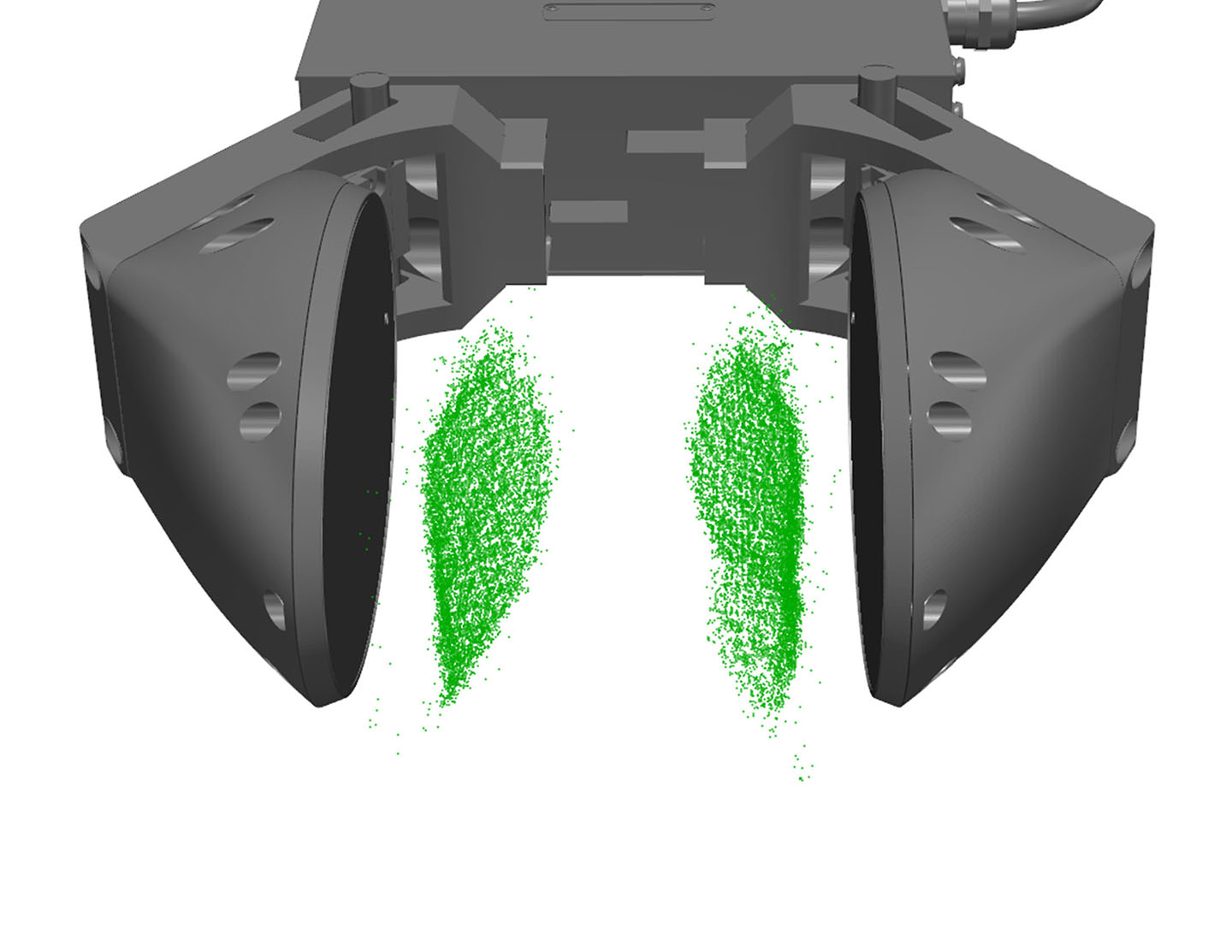}
  \caption{}
  \label{fig:proximity_pose_optimizer:3}
\end{subfigure}\hfil 
\begin{subfigure}{0.48\columnwidth}
  \includegraphics[width=\linewidth]{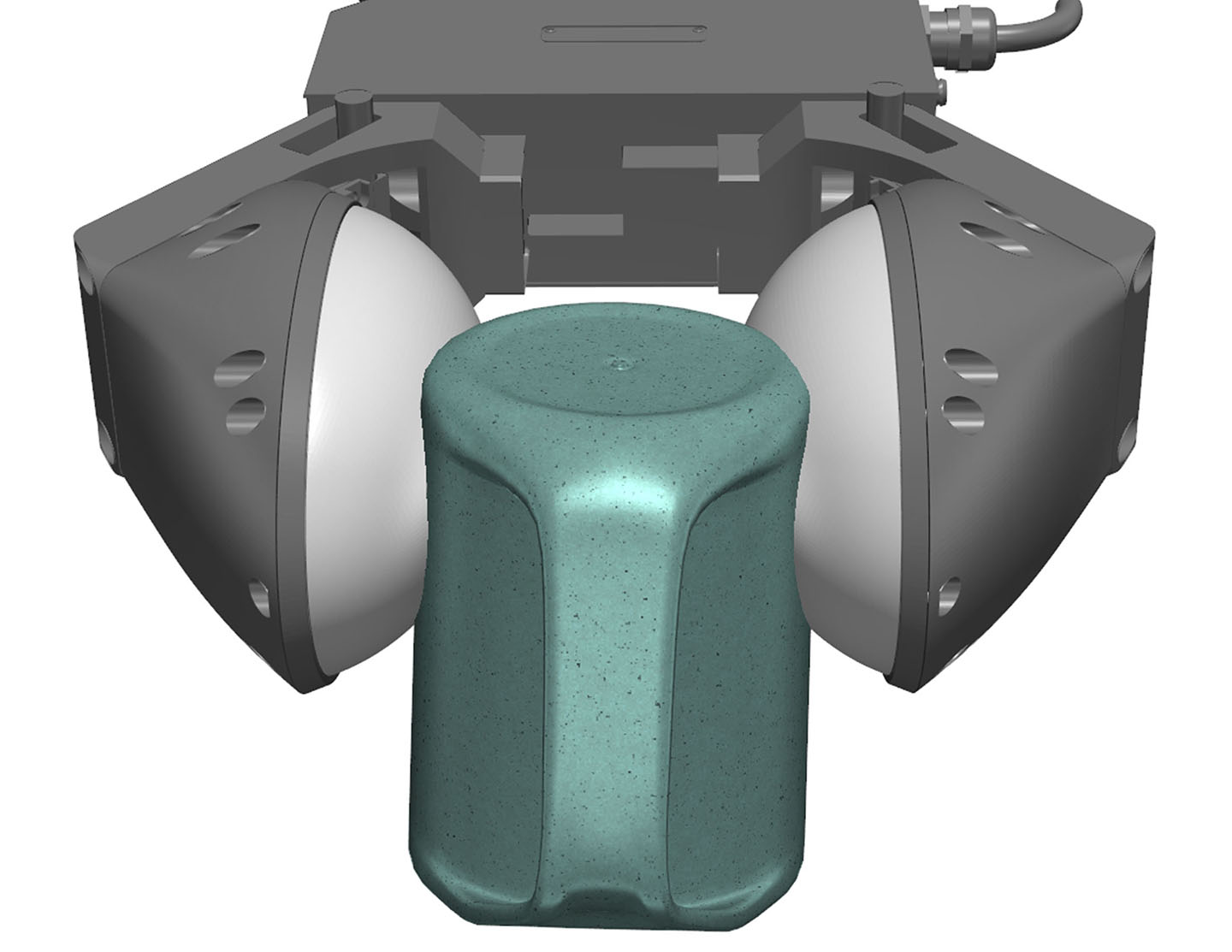}
  \caption{}
  \label{fig:proximity_pose_optimizer:4}
\end{subfigure}
\caption{The various stages of the in-hand pose estimation pipeline. (a) A plastic mug being grasped in the \softbubble gripper system. (b) The concatenated point-cloud produced from the depth images from each \softbubble sensor computed in the gripper frame. (c) The contact-patch filtered concatenated point-cloud (d) The estimated in-hand mug pose from the \emph{proximity} pose estimator.}
\label{fig:proximity_pose_optimizer}
\end{figure}

For \softbubbles we compute a concatenated grasp point-cloud $^Tp{^Q}_{{set}_i}$ from the two individual point-clouds by using gripper dimensions and the joint-encoder measurements from the parallel gripper. In practice, the speed and accuracy of this \emph{proximity} pose optimization were greatly enhanced by cropping this point-cloud to only include the contact-patch (similar to our earlier work \cite{kuppuswamy2019fast}). For the results shown in this paper, since we only test shapes that are convex, we used a more naive contact patch filter that utilized a difference image generated with respect to a reference configuration of the bubbles when they are free from contact. The various stages of this entire pipeline can be seen in Fig. \ref{fig:proximity_pose_optimizer}. In practice, the contact patch point-cloud size is typically around $10K$ points; the pose-estimation computation frequency tends to be around $0.5 Hz - 1 Hz$ on a standard multicore processor system using parallelization of computation of Eq. \ref{eq:proximity_pose_optimization}.

\section{Shear deformation estimation}
\label{sec:shear_deformation_estimation}
Incorporating a pseudorandom dot pattern on the interior of the bubble surface (Fig. \ref{fig:shear_patterns}) allows us to extract tangential displacement in the plane orthogonal to the applied gripper force ($P_{membrane}$). 

As shown in Sec. \ref{sec:mech_design}, the image sensor providing IR images is angled to look at the membrane. We extract a vector field using dense optical flow via the Gunner Farneback’s algorithm \cite{Farneback2003}. Since tracking algorithms can have difficulties handling sudden changes, we instead focus on the \emph{relative} shear-displacement problem of comparing two images: a reference $I_{t=0}$ and a later image $I_{t=n}$. The reference image is taken after the gripper has achieved a stable grasp of the target object. The second $I_{t=n}$ image is the latest from the camera. The output is vector $\Vec{V}$ for each pixel of the image that indicates an estimate of the displacement of that patch in pixels.
To extract the tangential displacement vector $\Vec{s}$, we sum the vector field on the image by $\Vec{s} = \sum_{i=0, \Vec{v}_{i} \in \Vec{V}}^{n} \Vec{v}_{i}$, where
$\Vec{s}$ is the tangential displacement, $v_{i}$ is the vector value at pixel $i$, $n$ is the pixel number of input image. The output relative shear displacement is then represented by the normalized form $\Vec{s} / \|\Vec{s}\|$. 



\section{Tactile object classification}
\begin{figure}[t]
\centering 
\begin{subfigure}{\columnwidth}
  \includegraphics[width=\linewidth]{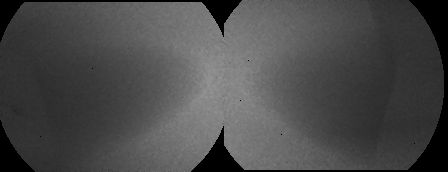}
\end{subfigure}\vfil 

\medskip

\begin{subfigure}{\columnwidth}
  \includegraphics[width=\linewidth]{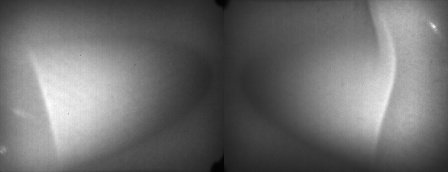}
\end{subfigure}\hfil 
\caption{An instance of an automatically labelled training sample for the plastic mug; Top: the concatenated depth image; Bottom: the concatenated IR image.}
  \label{fig:classifier_training}
\end{figure}

As described above, the ToF sensors in the bubbles provide both depth and IR over independent channels. We implemented a classifier that can utilize either of these streams from both cameras. Each camera's resolution is $224 \times 176$. Only stable grasps are used for training and inference - we compute if a grasp is stable using thresholds on the bubble pressure differential as well as the finger velocity. This eliminates confounding issues like image blur. The images from both bubbles are then concatenated (as in Fig. \ref{fig:classifier_training}) and down-sampled into a 224 x 224 image. ResNet18 was used as the network architecture and both training and inference pipelines were implemented with $pytorch$.

\subsection{Automatic Labelling Pipeline}
In contrast to our previous classifier \cite{Alspach2019}, we observed that when using images from two \softbubbles to classify objects during manipulation, it is important to collect training images of a sufficient diversity of valid stable grasps per class. Further, any unique geometric features of an object (e.g. handle of a mug, cap of a bottle) need to be sufficiently represented in the training data. Initial training attempts using human-generated grasp configurations resulted in insufficient samples of object geometries in poses the robot might grasp, resulting in poor classifier performance when tested on the robot. To mitigate these issues, we built an automatic labeling pipeline. One or more objects were placed into the robot's workspace and the robot was commanded to repeatedly pick up objects using the same grasp generator that is used in our object sorting control pipeline (antipodal grasps on the object computed using an external depth sensor \cite{ten2017grasp}) and then drop them randomly. Once a stable grasp is obtained, collecting about one minute of data (600 training samples) for each object was found to be sufficient for training. 

This pipeline can generate $50$ grasps ($30K$ training samples) autonomously within $1.5$ hours. We trained on three classes of objects (see Fig. \ref{fig:dishloading-classifier}). The training converged in about 45 epochs using stochastic gradient descent with a validation accuracy of $\ge 99\%$ (validated on a separate random samples of $30\%$ of the labelled data). For both training and inference we utilized a dual GPU Intel Xeon workstation. More rigorous analysis of the number of grasps required are ongoing and out of the scope of this paper.

\section{Experiments}
To demonstrate robust manipulation we deployed the \softbubble gripper and the presented tactile perception techniques in a cluttered indoor scenario. The experimental setup consisted of a fixed-base manipulator in a realistic kitchen setup with counters, a sink, and a dishwasher. The setup and the various tasks that we demonstrated can be seen in this \href{https://youtu.be/G_wBsbQyBfc}{video}.

\subsection{Robot and Task Setup}
\label{subsec:setup}
The bubble fingers were mounted on a parallel Schunk WSG 50 gripper attached to a KUKA iiwa arm. A set of six Intel RealSense D415 cameras were employed for external (visual) perception of the sink-scene; three cameras pointed into the sink at a downward angle, an additional two cameras facing the counter-top and one final camera tracking the three racks (using fiducial markers) and the door of the dishwasher. The following objects were used: plastic mugs, plastic PET bottles (acting as refuse in the sink), and wine glasses. An OptiTrack motion capture system was used for validating the in-hand pose estimation.

Depth images from the RealSense cameras were fused to generate a combined point cloud of the scene within the sink. Grasp poses were identified on objects either by an anti-podal grasp generation pipeline \cite{ten2017grasp} or by a naive blob centroid location. 

Motion planning was implemented using a combination of sampling-based and optimization-based techniques \cite{snoptIK}. The robot's actions were scripted in the form of a set of parameterized closed-loop \emph{primitives}. Real-time control ran over Ethernet/IP using KUKA's FRI interface, and the software framework used a mixture of tools written in C++ and Python. The Drake simulation and control toolbox \cite{bib:drake} was used for the kinematic computations within the control and planning stack. The collision detection in planning and control used the FCL collision library (through Drake). All of the perception, planning, and control software was running on one Intel Xeon workstation under Ubuntu 18.04.

\begin{figure}[ht!]
    \centering 
\begin{subfigure}{0.48\columnwidth}
  \includegraphics[width=\linewidth]{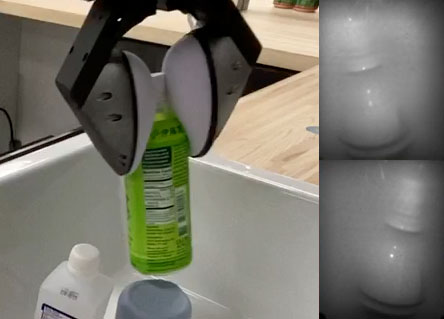}
  
  \label{fig:dishloading-classifier:1}
\end{subfigure}\hfil 
\begin{subfigure}{0.48\columnwidth}
  \includegraphics[width=\linewidth]{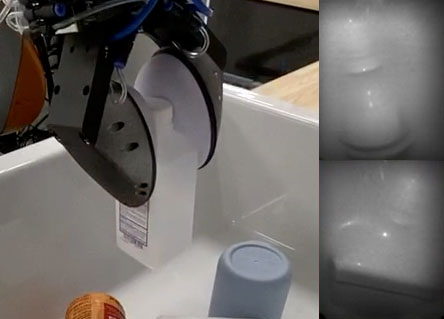}
 
  \label{fig:dishloading-classifier:2}
\end{subfigure}\hfil 

\medskip

\begin{subfigure}{0.48\columnwidth}
  \includegraphics[width=\linewidth]{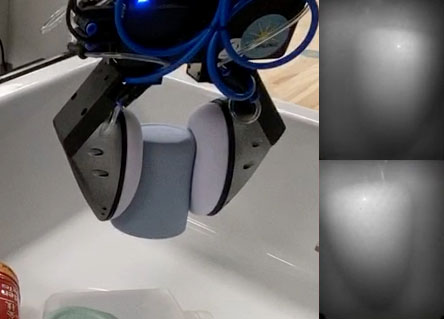}

  \label{fig:dishloading-classifier:3}
\end{subfigure}\hfil 
\begin{subfigure}{0.48\columnwidth}
  \includegraphics[width=\linewidth]{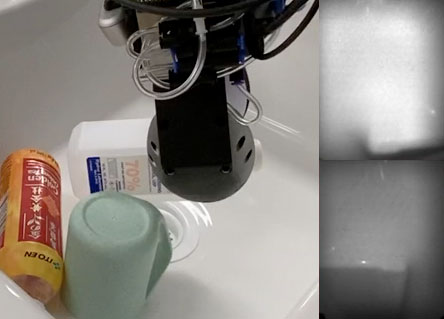}

  \label{fig:dishloading-classifier:4}
\end{subfigure}\hfil 

\caption{Some of the grasped objects from the dish loading task and their corresponding \softbubble IR images. In each of these cases, the object class was correctly identified with a probability $p>0.95$; (a) plastic PET tea-bottle, (b) square form-factor plastic alcohol bottle, (c) plastic mug, and (d) an alternative grasp utilized on a bottle similar to that in (b). }
\label{fig:dishloading-classifier}
\end{figure}


\subsection{Task 1: Dish loading with recycling separation}
\label{sec:dish_loading_task}
In this task the robot autonomously reaches into the sink, grasps objects, and classifies them as either a valid dish (e.g. mugs) or recycling (e.g. PET bottles). The limit on the objects selected was due to our existing dish-loading pipeline, not fundamental limitations of \softbubbles. Valid dishes must be placed within the dishwasher and recycling dumped in a bin. screen

The images seen in Fig. \ref{fig:dishloading-classifier} depict some of the different classes of objects that we tested on the dish-loading setup during the post-grasp tactile classification stage.


\subsection{Task 2: Pose Estimation Under Antagonism}
\label{sec:pose_antagonism}

We analyzed the performance of the \emph{proximity} pose optimizer for cylindrical mugs by equipping the manipuland with active motion capture markers. For an arbitrarily chosen in-hand pose, the pose estimates computed were compared against the motion capture output. Since the \emph{proximity} field we use here is that of a cylinder, the pose-estimates can be considered reliable only for contact with cylindrical surfaces. The peak range of the offset in the seed from the ground-truth pose, which still exhibited stable convergence, was observed to be $\pm 30$ degrees in pitch and roll directions. We used either the pre-grasp pose estimated by our vision system (when available) or seeded it with a cardinal pose. Once a stable grasp was achieved, we initialized our pose optimization with the previous estimated pose for continuous tracking.

\begin{figure}[t]
    \centering 
\begin{subfigure}{.48\columnwidth}
  \includegraphics[width=\linewidth]{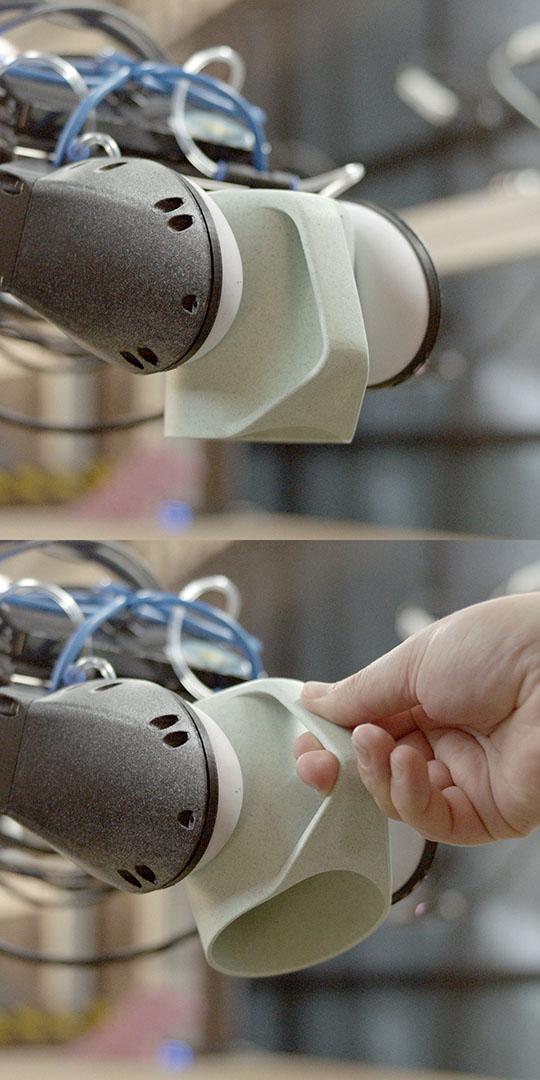}
\end{subfigure}\hfil 
\begin{subfigure}{.48\columnwidth}
  \includegraphics[width=\linewidth]{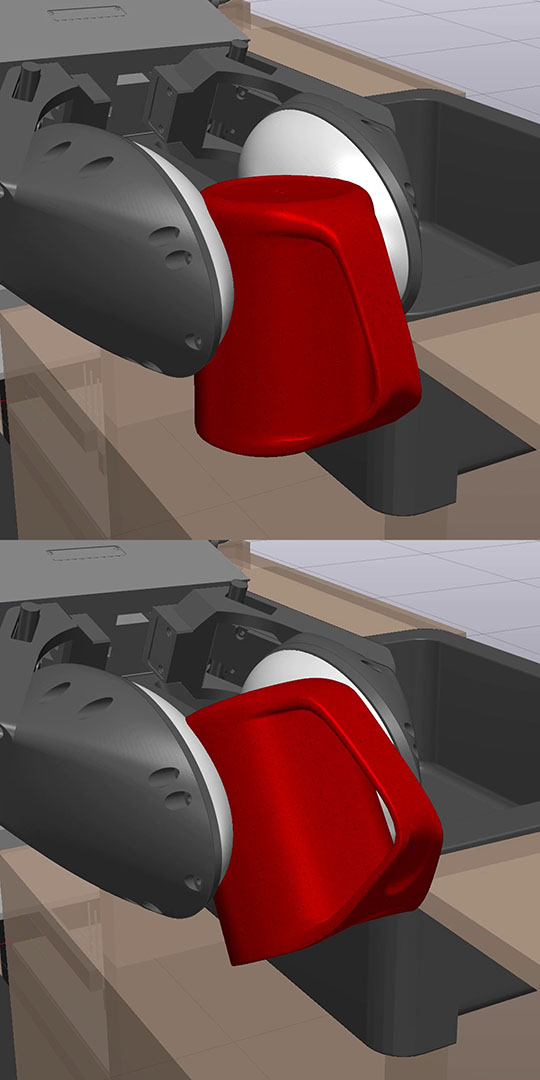}
\end{subfigure}\hfil 

\caption{Mugs grasped in two different poses and their corresponding in-hand pose-estimates; Top: An upside down mug; Bottom: Grasped mug being perturbed. The pose estimator runs at 1-2 Hz and can track these kinds of perturbations.}
\label{fig:mug-pose-perturbation}
\end{figure}

\subsection{Task 3: Shear-based Manipuland Release and Handover}
\label{sec:handover}

\begin{figure*}[!ht]
\vspace{6pt}
    \centering 
\begin{subfigure}{0.325\textwidth}
  \includegraphics[width=\linewidth]{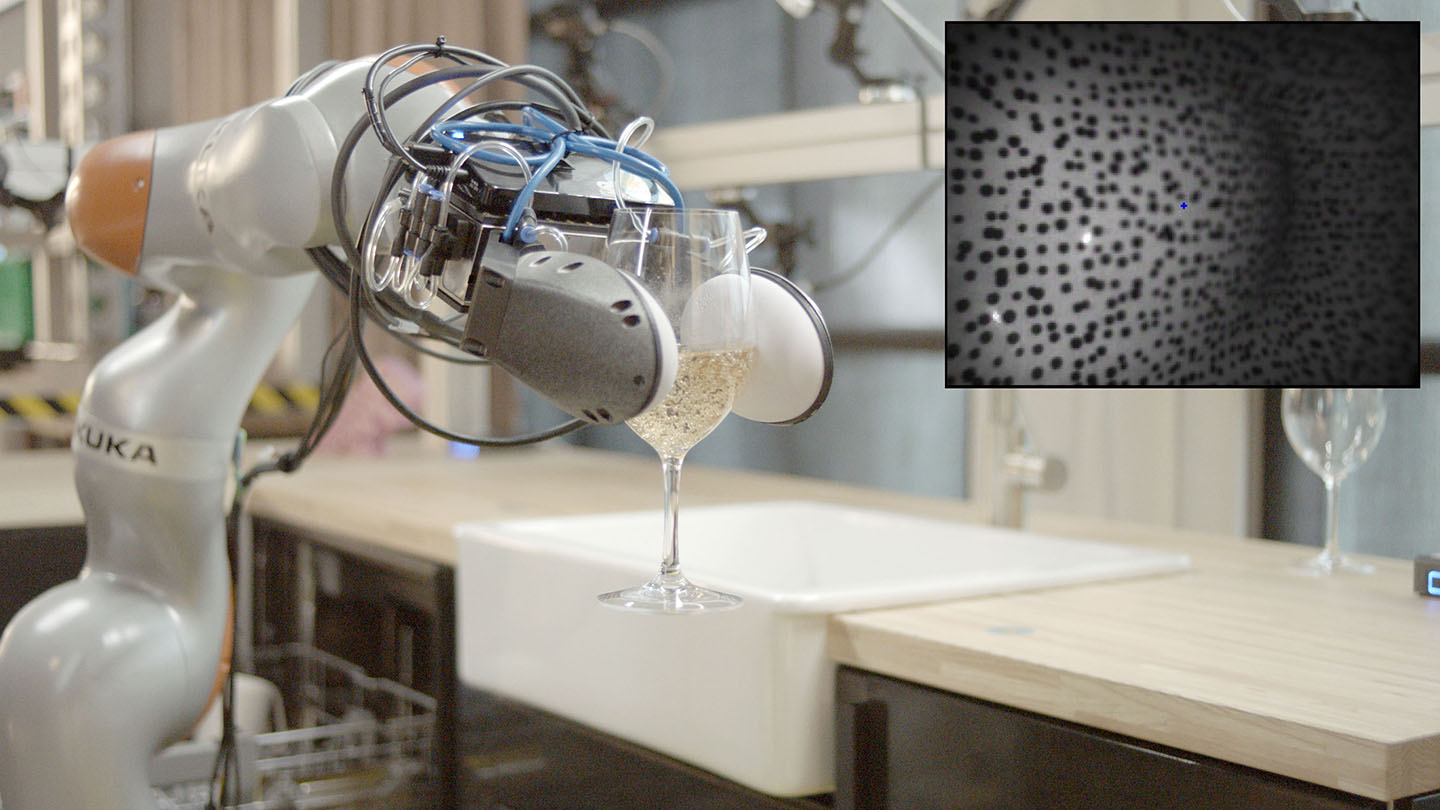}
  
  \label{fig:handover:1}
\end{subfigure}\hfil 
\begin{subfigure}{0.325\textwidth}
  \includegraphics[width=\linewidth]{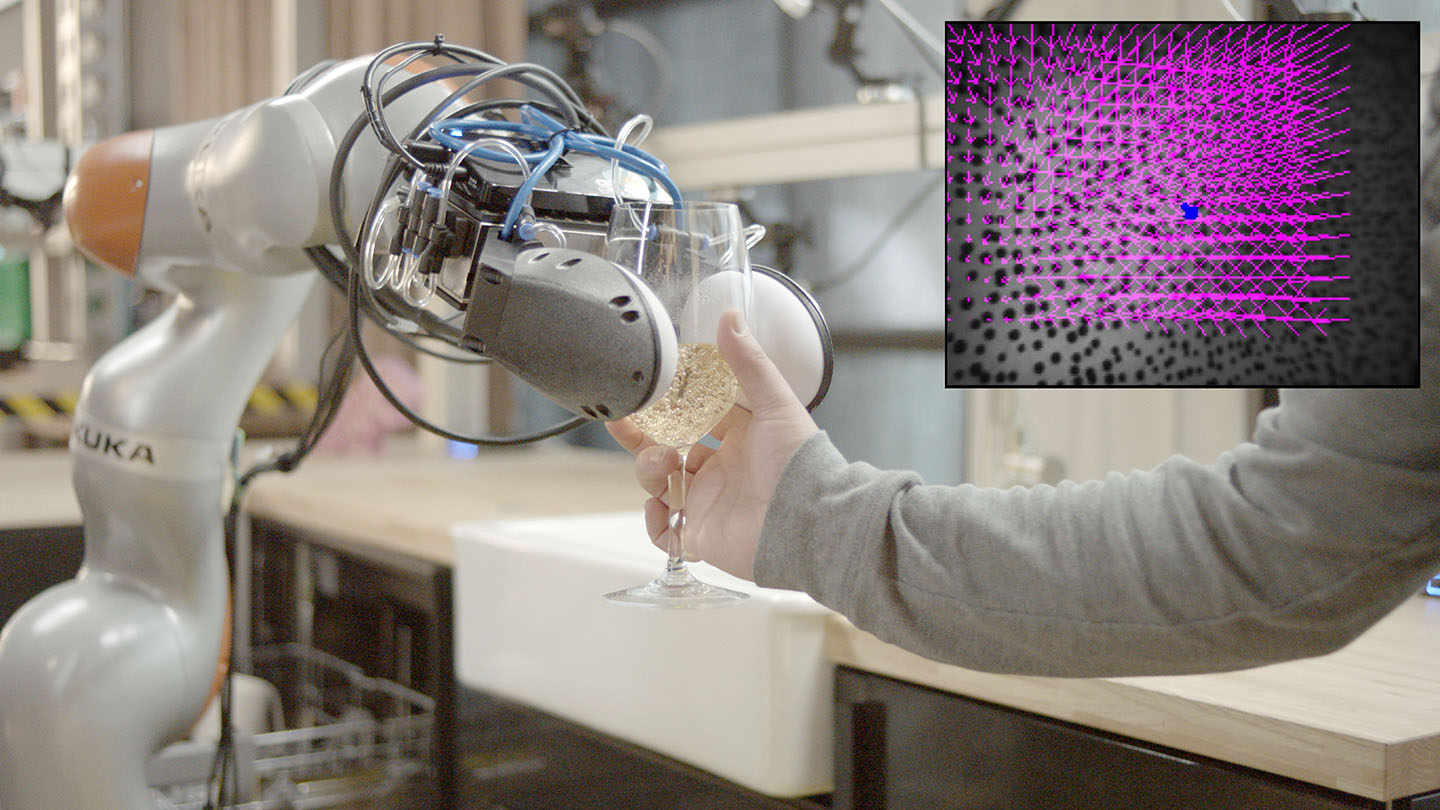}
  
  \label{fig:handover:2}
\end{subfigure}\hfil 
\begin{subfigure}{0.325\textwidth}
  \includegraphics[width=\linewidth]{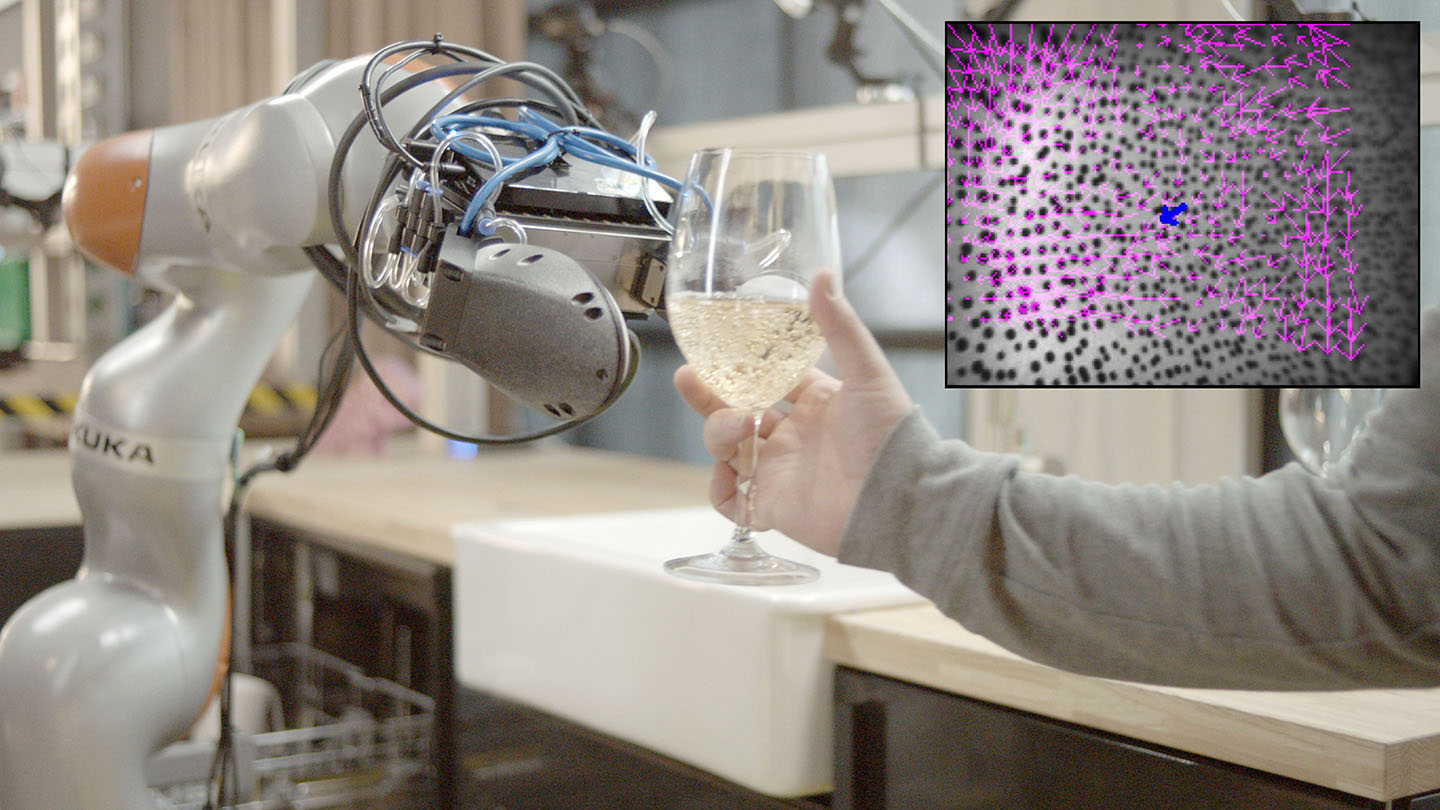}
  
  \label{fig:handover:3}
\end{subfigure}\hfil 
\caption{Video frames from a shear-based manipuland handover task with a liquid filled fragile manipuland (wine-glass). The shear displacement estimator's display output (computed from one of the \softbubble fingers) is overlaid on the top right of each panel: (left) Before handover, showing the nominal configuration. From this point on, shear is computed relative to this configuration. (center) At the onset of handover, when a human exerts forces to take the manipuland from the robot, relative shear displacement increases until a threshold on the norm of the magnitude. (right) Condition immediately after release; there is residual relative shear seen with respect to the pre-handover condition.}
\label{fig:handover}
\end{figure*}

We implemented a closed-loop controller that monitors the shear forces on a manipuland relative to an initial stable grasp. By comparing it to a threshold we implemented automated placing or handover of the manipuland. This was demonstrated with two tasks. In the first we hand a fragile object (wine glass with liquid) to a human as seen in Fig. \ref{fig:handover}. The second is an autonomous blind stacking of wine glasses where the robot has no notion of the geometry of the objects or the stacking surface. This latter task can be seen in this    \href{https://youtu.be/G_wBsbQyBfc}{video link}.

The thresholds used are dependent on the post-grasp contact patch size and geometry. Future work will involve using the contact patch estimator and pressure sensor data to improve shear sensing accuracy for more diverse objects.


\section{Conclusions and Discussion}
\label{sec:conclusions_and_discussion}
In this paper, we present the \Softbubble gripper system, which combines a highly compliant gripping surface and visuotactile sensing in the form factor of a parallel gripper. The tactile sensing capability of the gripper enables multiple forms of perception; in multiple real-world tasks, we demonstrated in-hand pose-estimation, shear-deformation estimation and tactile classification, within a robust manipulation pipeline. 

As seen in our results, the combination of \softbubbles and tactile perception can be valuable for solving manipulation problems in cluttered environments. The \softbubble fingers are affordable, compliant, easy to construct and mechanically robust. In addition, the perception methods presented are computationally efficient enough to enable closed-loop, real-time control of complex tasks. The \emph{proximity} field method used in our pose estimation framework is a novel contribution that is promising for achieving tractable depth-based tracking.

\softbubbles still have room for improvement. Our current implementation is limited in size due to the minimum range of custom ToF sensors that are difficult to source, so we are working on customized ToF cameras designed specifically for short-range depth sensing.

Algorithmically, we are working on enhancements to both shear displacement estimation and pose tracking. Our current shear displacement tracking is based on tracking visual features with the IR channel. By incorporating the depth channel from the ToF camera we hope to have the \softbubble act more like a full low-cost force-torque sensor. Our pose estimator is being expanded to handle other geometric primitives (e.g. frustrums).

Lastly, we are investigating more challenging applications for the \softbubble technology, including tool use and enabling physical interaction between humans and robots.


\bibliographystyle{IEEEtran}
\bibliography{IEEEabrv,references} 

\end{document}